\documentclass{article}

\usepackage{microtype}
\usepackage{graphicx}
\usepackage{subfigure}
\usepackage{booktabs} 

\usepackage{amsmath}
\usepackage{amsfonts}
\usepackage{mathtools}

\mathtoolsset{showonlyrefs=true}

\usepackage{hyperref}

\usepackage[accepted]{icml2022}

\icmltitlerunning{Feature Space Particle Inference 
           for Neural Network Ensembles}

\begin{document}

\twocolumn[
\icmltitle{Feature Space Particle Inference 
           for Neural Network Ensembles}



\icmlsetsymbol{equal}{*}

\begin{icmlauthorlist}
\icmlauthor{Shingo Yashima}{itlab}
\icmlauthor{Teppei Suzuki}{itlab}
\icmlauthor{Kohta Ishikawa}{itlab}
\icmlauthor{Ikuro Sato}{itlab,titec}
\icmlauthor{Rei Kawakami}{itlab,titec}
\end{icmlauthorlist}

\icmlaffiliation{itlab}{Denso IT Laboratory Inc., Tokyo, Japan}
\icmlaffiliation{titec}{Tokyo Institute of Technology, Tokyo, Japan}

\icmlcorrespondingauthor{Shingo Yashima}{yashima.shingo@core.d-itlab.co.jp}

\icmlkeywords{Machine Learning, ICML}

\vskip 0.3in
]



\printAffiliationsAndNotice{}  

\begin{abstract}
Ensembles of deep neural networks demonstrate improved performance over single models. For enhancing the diversity of ensemble members while keeping their performance, particle-based inference methods offer a promising approach from a Bayesian perspective.
However, the best way to apply these methods to neural networks is still unclear:
seeking samples from the weight-space posterior suffers from inefficiency due to the over-parameterization issues, while seeking samples directly from the function-space posterior often results in serious underfitting. 
In this study, we propose optimizing particles in the feature space where the activation of a specific intermediate layer lies to address the above-mentioned difficulties.
Our method encourages each member to capture distinct features, which is expected to improve ensemble prediction robustness. 
Extensive evaluation on real-world datasets shows that our model significantly outperforms the gold-standard Deep Ensembles on various metrics, including accuracy, calibration, and robustness.
\end{abstract}

\section{Introduction}
Averaging predictions of multiple trained models has been a very popular technique in machine learning because it can significantly improve the generalization ability of a prediction system. In particular, ensemble methods of neural networks have recently achieved significant success in terms of predictive performance \citep{lakshminarayanan2017simple}, uncertainty estimation \citep{ovadia2019}, and robustness to adversarial attacks \citep{pang2019improving} or perturbations \citep{hendrycks2019benchmarking}. 
Among many ensemble methods, Deep Ensembles \citep{lakshminarayanan2017simple}, which train each model independently from randomly initialized weights, has been the de facto approach with a decent performance and ease of implementation. However, it relies only on the randomness of initialization to generate different ensemble members and thus does not explicitly encourage diversity among them, which can cause redundancy in model averaging \cite{rame2021dice}.

More recently, particle-based variational inference methods \citep{liu2016stein} have provided a promising approach for composing better ensembles from a Bayesian perspective \citep{wang2019function, d2021repulsive}. These methods intend to approximate the Bayes posterior by the (pre-specified) finite number of models (called {\it particles}) using deterministic optimization. They have greater non-parametric flexibility than classical variational inferences \citep{blundell2015weight, gal2016dropout} and provide better efficiency than sampling-based Markov chain Monte Carlo (MCMC) methods \citep{neal1996bayesian, welling2011bayesian}. Notably, their optimizations take interactions between models into account using kernel functions and explicitly promote model diversity, unlike Deep Ensembles. 

These inferences are usually performed on the weight space of neural networks to approximate weight-space posterior. However, this is far from ideal due to the over-parameterized nature of recent neural networks. Such models have many local modes in the weight-space posterior that are distant from each other, yet corresponding to the same predictive function \citep{fort2019deep, entezari2021role}. Therefore, promoting diversity of weights does not necessarily result in diversity as a predictive function, and it can yield a degenerate ensemble when the number of models is limited. 

To circumvent this, Wang et al. \citeyearpar{wang2019function} proposed performing the inference on model outputs to obtain an approximation of the function-space posterior. They treat output logits on data points as inferred parameters and promote diversity on them.
Although they do not suffer from the above-mentioned degeneration issues, directly seeking a posterior on the output space of neural network functions often results in severe underfitting \citep{d2021repulsive}. Overall, none of these methods have shown significant improvements over naive Deep Ensembles in accuracy and calibration \citep{d2021stein}.

On the other hand, a recent theoretical study suggests that the critical component of the success of neural network ensembles is the {\it multi-view} structure of data \citep{allen2020towards}. They claimed that when data have multiple features that can be used to classify them correctly (which they call multi-view), and each member of an ensemble captures different features from each other, the prediction performance on the test data can be boosted by ensembling. This multi-view structure is typical in real-world data: for example,  the image of a male lion presented in \autoref{lion} can be classified correctly by looking either at its mane or face. Intuitively, if we ensemble models which look at different parts of a lion to classify, the robust prediction would be possible even on, for example, female lions without a mane. 

Leveraging these works, we hypothesize that promoting the diversity of feature extractors rather than predictive functions encourages each model to capture different data views, resulting in better performance of an ensemble. We expect that feature diversity allows the same predictive functions on easy data, which prevents underfitting, and yields different predictive functions on hard data, which increases robustness. To this end, we propose a framework for performing particle-based variational inference on the feature space of networks. Our contributions are summarized as follows:

\begin{figure}[t]
\vskip 0.2in
\begin{center}
\centerline{\includegraphics[width=\columnwidth]{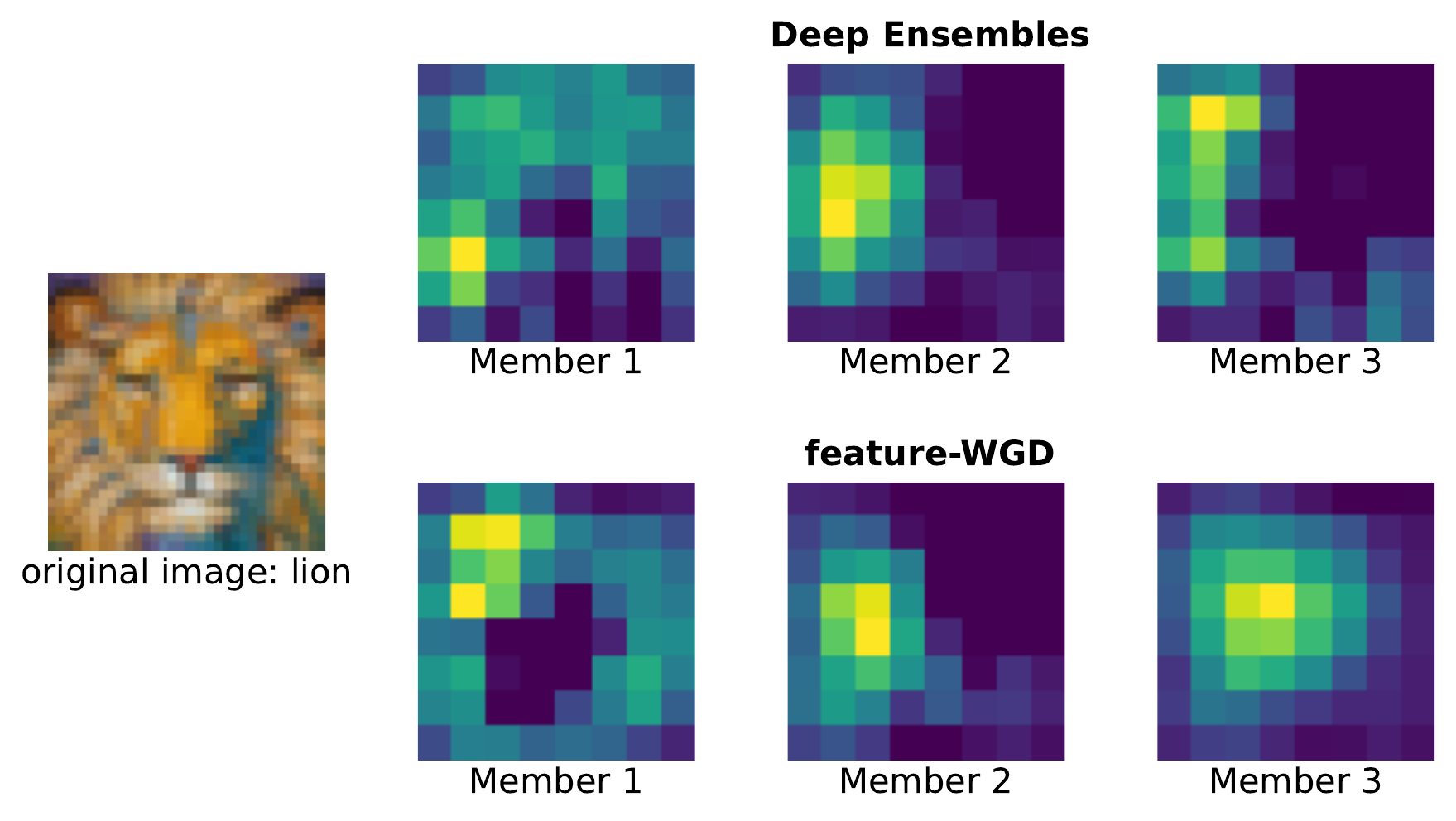}}
\caption{Class activation maps for an ensemble of WRN-28-2 on CIFAR-100 trained with Deep Ensembles (up) and the proposed method (bottom). The proposed method captures more diverse features (mane, face) than Deep Ensembles. Use Grad-CAM \citep{selvaraju2017grad} for visualization.}
\label{lion}
\end{center}
\vskip -0.2in
\end{figure}

\begin{itemize}
    \item We formalize particle-based inference methods on feature space of neural networks (\autoref{sec3}), which encourages each network to capture distinct features to classify data. In our construction, multiple feature extractors are connected to a common classification layer, so that the feature distribution from all feature extractors is adequately controlled with respect to the subsequent layer. It does not suffer from over-parameterization issues in weight-space inference, and it is hard to underfit, unlike function-space inference.
    \item We performed extensive experiments to evaluate the classification and calibration ability of the proposed approach on typical image datasets, including CIFAR and ImageNet (\autoref{experiments}). The results show the superiority of feature-space inference over weight or function-space inference. Moreover, the proposed approach consistently outperforms gold-standard Deep Ensembles in generalization ability and robustness.
\end{itemize}

\section{Background}

\label{background}

\subsection{Bayesian neural networks and Deep Ensembles}
In typical supervised deep learning, we consider a likelihood function $p(y|f(x;w))$ with a neural network $f$ parameterized by $w$. Then we maximize the likelihood of training data $\mathcal{D} = \left \{ (x_b, y_b) \right \}_b$ with respect to $w$ to obtain a network that adequately explains the data.
In Bayesian neural networks (BNNs), we intend to ensemble all-likely networks which explain training data $\mathcal{D}$ by drawing weights from the posterior distribution $p(w|\mathcal{D}) \propto p(w) \prod_{(x, y)\in \mathcal{D}} p(y|f(x; w))$, where $p(w)$ is a prior over weights. When making a prediction on a test point $x_*$, we marginalize the predictive functions over the posterior:
\begin{align}
    p(y_*|x_*, \mathcal{D}) = \int_\mathcal{W} p(y_* | f(x_*; w)) p(w|\mathcal{D}) \mathrm{d}w \label{predictive}
\end{align}
The exact posterior is generally intractable in the case of neural networks; thus, various approximation methods for BNN have been developed so far, including variational inference methods \citep{graves2011practical, blundell2015weight, gal2016dropout, wen2018flipout} and MCMC methods \citep{neal1996bayesian, welling2011bayesian}. With weights $\{w_i\}_{i=1}^n$ sampled from the (possibly approximated) posterior, the predictive distribution is obtained by Monte Carlo estimates of \eqref{predictive}:  $p(y_*|x_*, \mathcal{D}) \approx \frac{1}{n} \sum_{i=1}^n p(y_*|f(x_*;w_i))$.

On the other hand, Deep Ensembles \citep{lakshminarayanan2017simple} compose an ensemble of networks by initializing and training each network independently. Although they do not explicitly sample network parameters from particular distributions like BNNs, diverse networks which explain data can be obtained, because different initial values of each network weights result in different solutions in non-convex neural network training \citep{fort2019deep}.

From a practical viewpoint, it is known that BNNs are generally inferior to Deep Ensembles in terms of both accuracy and uncertainty estimation \citep{ashukha2019pitfalls, ovadia2019, gustafsson2020evaluating}. This is because the expressive power of the approximated posterior in practical BNNs is not sufficient to capture various modes in the complex weight-space posterior. As a result, each member in a BNN ensemble is less diverse as a predictive function than that in Deep Ensembles \citep{fort2019deep}. 
\subsection{Particle-based variational inference for BNNs}
Recently, particle-based variational inference methods \citep{liu2016stein, chen2018unified} have received attention for developing better ensemble methods in BNNs.
These methods seek to approximate the target posterior $p(w|\mathcal{D})$ with the pre-specified number of particles $\{w_i\}_{i=1}^n$ by transporting them using deterministic optimization. Its non-parametric flexibility allows it to capture more complex posterior than traditional variational inference methods. In addition, the interaction between particles in optimization enables a more particle-efficient approximation of the posterior than sampling-based MCMC.

In typical particle-based inference methods, such as Stein variational gradient descent (SVGD) \citep{liu2016stein}, an update direction $v$ for each particle $\{w_i\}_{i=1}^n$ is described in the following form:
\begin{align}
    v(w_i) = \sum_{j=1}^n \beta_{ij} \nabla \log p(w_j|\mathcal{D}) + \gamma_{ij} \nabla_{w_j} k(w_i, w_j)
\end{align}
where $k$ is a positive definite kernel and $\beta_{ij}$ and $\gamma_{ij}$ are scalars that depend on $\{w_i\}_{i=1}^n$. 
As we can see, the update direction $v$ consists of two parts: the {\it driving term}, which pushes particles towards high-density regions in the posterior, and the {\it repulsive term}, which prevents particles from collapsing into a single MAP estimate. Thus, their training procedure can be viewed as Deep Ensembles with repulsive forces between members \cite{d2021repulsive}. They explicitly promote diversity through member interactions and, therefore, can be expected to produce a better performing ensemble than naive Deep Ensembles.

However, when applied to over-parameterized models such as neural networks, they can produce degenerate ensemble members when the number of particles is limited, whereas a consistency to the true posterior is guaranteed in many particle limits \citep{liu2017stein}. That is, because different weights far from each other can map to the same function in such over-parameterized models, the repulsive term on weights does not effectively promote diversity as a predictive function.

Regarding this, Wang et al. \citeyearpar{wang2019function} proposed to directly seek a functional posterior $p(f|\mathcal{D}) \propto p(f) \prod_{(x, y)\in \mathcal{D}} p(y|f(x))$ by performing inference on the output logits of a network evaluated on data points. However, it often shows severe underfitting in real-world image classification tasks \cite{d2021stein}. The reason behind this is still not apparent, but one possible explanation is the dangers of directly promoting diversity in logit space. In typical image datasets such as CIFAR, a label is almost deterministically produced for a given image (i.e., there is low label noise \citep{tsybakov2004optimal}). Thus, there is no room for diversity in the logit space for most images in such datasets. 
More recently, D'Angelo et al. \citeyearpar{d2021stein} proposed a hybrid use of weight-space and function-space update rules. They showed an improved performance using additional stochasticity on gradients \citep{gallego2018stochastic}, but Deep Ensembles still perform the best in the deterministic setting in terms of accuracy and negative log-likelihood.


\section{Particle-based Inference on Feature Space}
\label{sec3}
In the following, we present a framework for feature space particle-based inference. 
Its advantages over weight or function-space inference are summarized as follows:
\begin{itemize}
    \item Given a shared classifier on the top of independent feature extractors as described in follows, a posterior of feature extractors is considered to be relatively simple and hard to suffer from the over-parameterization problem observed in weight-space inference.
    \item It does not directly promote diversity on output logits, even allowing the same prediction results for easy data. This is expected to prevent the underfitting observed in the function-space inference.
    \item By encouraging each feature extractor to capture different features to classify the same input data, it can exploit the multi-view structure of data, which is considered a critical component of the performance gain by ensembling.
\end{itemize}

We begin by introducing a specific form of the inference method used in this study and then show how to perform inference on feature extractors.
Finally, we discuss the prior selection of feature extractors and the computational efficiency of our algorithm.

\subsection{Wasserstein gradient descent}
There are various formulations of particle-based variational inference, depending on how variational approximation and discretization are applied to derive finite-particle update rules.
This section introduces a specific form of the inference method used in this study, named Wasserstein gradient descent (WGD) \cite{liu2019understaning, wang2021projected, d2021repulsive}. Note that the choice of an inference method is independent of the space in which the inference is performed, and we denote the variables to be inferred as $w \in \mathcal{W}$ here.

 Given the target posterior distribution $p(\cdot|\mathcal{D})$, our final objective is to minimize the following KL divergence with respect to the current particle distribution $q$:
\begin{equation}
    \mathrm{KL}_{p(\cdot|\mathcal{D})}(q) = \int_{\mathcal{W}} (\log p(w|\mathcal{D}) - \log q(w))p(w| \mathcal{D}) \mathrm{d}w. 
\end{equation}
Particle-based inference can be formulated as a gradient descent optimization of the above functional in the Wasserstein space $\mathcal{P}_2(\mathcal{W})$ equipped with the well-known Wasserstein distance $W_2$ \citep{ambrosio2008gradient, villani2009optimal}.
Specifically, its Wasserstein gradient flow  $\{(q_t)_t\}$, which is roughly the family of steepest descending curves for $\mathrm{KL}_{p(\cdot|\mathcal{D})}$ in $\mathcal{P}_2(\mathcal{W})$, has its tangent vector $v_t$ at any $t$ being:
\begin{equation}
    v_t(w) = \nabla \log p(w|\mathcal{D}) - \nabla \log q_t(w) \label{1}
\end{equation}
whenever $q_t$ is absolutely continuous \citep{liu2019understaning}. Intuitively, $v_t(w)$ represents the direction in which the probability mass on the point $w$ of $q_t$ should be moved in order to bring $q_t$ close to $p$. 

Although the first term of \eqref{1} can be calculated using an unnormalized posterior, we do not have access to the analytical form of the second term $\nabla \log q_t$ when performing gradient descent along with $v_t$ using finite particles $\{w_i^t\}_i$. In WGD, we consider using kernel density estimation (KDE) to approximate $q_t$ with $\{w_i^t\}_i$: $q_t(w) \propto {\sum_{j=1}^n k(w, w_j^t)}$. Here $k$ is a given positive definite kernel function such as RBF kernel. Then, an approximation of the second term in \eqref{1} is given by
\begin{equation}
    \nabla \log q_t(w_i^t) \approx \frac{\sum_{j=1}^n \nabla_{w_i^t} k(w_i^t, w_j^t)}{\sum_{j=1}^n k(w_i^t, w_j^t)}.
\end{equation}
Using the above formula, the update rule of particle $w_i^t$ is obtained as follows:
\begin{align}
    &w_i^{t+1} = w_i^t + \alpha_t v_t(w_i^t)\\
    &\approx w_i^t + \alpha_t \left(\nabla \log p(w_i^t|\mathcal{D})  - \frac{\sum_{j=1}^n \nabla_{w_i^t} k(w_i^t, w_j^t)}{\sum_{j=1}^n k(w_i^t, w_j^t)}\right),\label{2}
\end{align}
where $\alpha_t>0$ is a step size parameter for iteration $t$.
Unlike SVGD, the update rule of WGD does not include the averaging of the gradient of the log posterior between particles, which is known to be harmful in high-dimensional settings such as neural networks \cite{d2021repulsive}. We note that, although several studies proposed more particle-efficient update rules than WGD \citep{li2018gradient, shi2018spectral}, we do not go deep into a choice of inference algorithm itself and use WGD as a simple baseline.

\subsection{WGD on feature space}
Let $h(\cdot;w)$ be a feature extractor and $c(\cdot ;\theta)$ be a classifier of the neural network $f$ parametrized by $w$ and $\theta$, respectively: $f(\cdot; w, \theta) = c(\cdot ;\theta) \circ h(\cdot;w)$. In typical image classification networks like ResNet \citep{he2016deep}, $c$ corresponds to a final linear layer and $h$ corresponds to a whole network before that.
We consider each ensemble member to have an independent feature extractor $h(\cdot; w_i)$ and a shared classifier $c(\cdot; \theta)$; therefore, $i$-th member of an ensemble is written as  $f(\cdot;w_i, \theta) = c(\cdot; \theta) \circ h(\cdot; w_i)$. The classifier sharing is essential for our formulation: by doing this, the output space of each feature extractor does not suffer from permutations of the subsequent classifier and thus is expected to share same semantic information, which enables performing particle inference on feature extractors as described below.

In this study, we consider a shared classifier $c(\cdot; \theta)$ as a deterministic function, whose  parameters are not inferred  in particle optimization. Thus, a posterior distribution that we seek to approximate by particles is given by
\begin{align}
    p(h|\mathcal{D}) &\propto p(h) p(\mathcal{D} | c(\cdot; \theta) \circ h) \\
    &=p(h) \prod_{(x, y)\in \mathcal{D}} p\left(y \middle| c (h (x);\theta)\right).
\end{align}
However, obtaining a functional posterior of $h$ is neither tractable nor practical in the case of neural networks. Following \citep{wang2019function, d2021repulsive}, we instead perform an inference on feature values evaluated at training points $\mathcal{X}=\{x_b\}_b$: $\mathbf{h} = \mathrm{vec}\left(\{h(x)\}_{x \in \mathcal{X}}\right)$.
Plugging into the update rule of WGD \eqref{2}, the update direction of $\mathbf{h}_i^t$ ($i$-th particle on iteration $t$) is written as
\begin{align}
    v_t(\mathbf{h}_i^t) = \nabla_{\mathbf{h}_i^t} \log p(\mathbf{h}_i^t|\mathcal{D})
    - \frac{\sum_{j=1}^n \nabla_{\mathbf{h}_i^t} k(\mathbf{h}_i^t, \mathbf{h}_j^t)}{\sum_{j=1}^n k(\mathbf{h}_i^t, \mathbf{h}_j^t)}. \label{eq8}
\end{align}
The second term is the repulsive force, which encourages the features of each member to be different from each other, whereas the first term promotes them to correctly classify data. The first term, the log-gradient of the posterior, is decomposed as follows:
\begin{align}
\label{eq3}
   \nabla_{\mathbf{h}_i^t}  \log p(\mathbf{h}_i^t | \mathcal{D}) =\nabla_{\mathbf{h}_i^t} \log p(\mathbf{h}_i^t) + \nabla_{\mathbf{h}_i^t} \log p(\mathcal{D} | \mathbf{h}_i^t).
\end{align}
The first term is a log-gradient of the prior, and the prior choice is discussed afterward.
The second term corresponds to a log-gradient of the data likelihood with respect to the feature values, given a shared classifier:
\begin{align}
    \nabla_{\mathbf{h}_i^t} \log p(\mathcal{D} | \mathbf{h}_i^t) = \nabla_{\mathbf{h}_i^t} \sum_{(x, y)\in \mathcal{D}} \log p(y | c (h (x; w_i^t);\theta)).
\end{align}
In the training procedure, we transport the feature extractors $\{\mathbf{h}_i\}_{i=1}^n$ by updating the weights $\{w_i\}_{i=1}^n$. An update rule of weights can be obtained by projecting the functional update \eqref{eq8} back to the weight space using a Jacobian:
\begin{align}
    w_i^{t+1} = w_i^{t} + \frac{\alpha_t}{|\mathcal{D}|} \left(\frac{\partial \mathbf{h}_i^t}{\partial w_i^t}\right)^\top v_t(\mathbf{h}_i^t). \label{eq6}
\end{align}
This update can be implemented using standard back-propagation. In addition to the particle optimization in the feature space, we simultaneously update the shared classifier $c(\cdot; \theta)$ by maximizing the average of log-likelihood over the particles and data:
\begin{align}
    \theta^{t+1} = \theta^t + \frac{\alpha_t}{n|\mathcal{D}|} \sum_{i=1}^n  \sum_{(x, y)\in \mathcal{D}} \nabla_{\theta^t} \log p\left(y \middle| c (h (x; w_i^t);\theta^t)\right). \label{eq7}
\end{align}
In practice, the update \eqref{eq6} and \eqref{eq7} can be performed in a mini-batch manner by modifying the feature values $\mathbf{h}_i^t$ to that evaluated on a current mini-batch. In addition, weight decay is applied to these parameter updates to prevent overfitting.
For making a prediction on a test point $x_*$ after training, we approximate the predictive distribution \eqref{predictive} using the obtained weights $\{w_i\}_{i=1}^n$ and $\theta$:
\begin{align}
    p(y_*| x_*, \mathcal{D}) \approx \frac{1}{n} \sum_{i=1}^n p(y_* | f(x_*; w_i, \theta)).
\end{align}

\textbf{Projection in the repulsive term.} Although the over-parameterization issue is circumvented, the inferred parameter $\mathbf{h}$ is generally high-dimensional, which causes inefficient sampling from the posterior. In particular, the repulsive term of the update rule \eqref{eq8} involves KDE, which is known to suffer from the curse of dimensionality \citep{scott1991feasibility}. Thus, inspired by \citep{wang2021projected, chen2020projected}, we consider estimating the density in a low-dimensional subspace in which the likelihood of data changes significantly. To find such a subspace, we use the gradient information of the log-likelihood by defining a matrix $H_t$ as 
\begin{align}
    H_t = \frac{1}{n} \sum_{i=1}^n \nabla_{\mathbf{h}_i^t} \log p(\mathcal{D} | \mathbf{h}_i^t) \left(\nabla_{\mathbf{h}_i^t} \log p(\mathcal{D} | \mathbf{h}_i^t)\right)^\top \label{eqht}
\end{align}
and use a $r$-dimensional subspace spanned by $\Psi_r = (\psi_1, \ldots, \psi_r)$, where $\psi_i$ is the $i$-th dominant eigenvector of $H_t$ and $r \leq n$. Denoting $\mathbf{z}_i^t = \Psi_r^\top \mathbf{h}_i^t \in \mathbb{R}^r$, the projected version of the repulsive term in \eqref{eq8} is obtained as
\begin{align}
   \Psi_r \frac{\sum_{j=1}^n \nabla_{\mathbf{z}_i^t} k( \mathbf{z}_i^t, \mathbf{z}_j^t)}{\sum_{j=1}^n k(\mathbf{z}_i^t, \mathbf{z}_j^t)}. \label{eq9}
\end{align}
Roughly speaking, we evaluate the repulsive term only on feature elements that significantly affect the prediction results. Note that in \citep{wang2021projected, chen2020projected}, the entire update rule including the driving term is projected onto a subspace, but we only project the repulsive (KDE) term because we found that yields more stable training on neural networks.

In addition, this projection can mitigate the risk of promoting a “useless” diversity of features. As in single model training, different feature dimensions of a trained ensemble may carry similar semantic information. As a result, the obtained diversity on such feature dimensions may not represent true semantic diversity. However, in such a case, there would be a subspace that does not influence prediction results. For example, when subsequent classifier weights on two feature elements are perfectly identical, increasing one feature value and decreasing the other does not affect the output logits. Such a redundant subspace is likely to be removed by the projection because we compose subspace by adopting directions that strongly influence prediction results, as in \eqref{eqht}.

We denote the  WGD implemented in each space as \{weight, function, feature\}-WGD, respectively. The entire inference procedure is summarized in \autoref{algorithm}.

\subsection{Prior choice}
\label{prior}
How to impose a functional prior on the feature extractor $h$ (or $\mathbf{h}$) is a major concern. It can be defined either as a push-forward measure of the weight-space prior or a stochastic process (e.g., Gaussian process (GP)) configured regardless of the parametric form of a network. The former approach can be implemented using an empirical approximation using weight samples \cite{wang2019function}, or a gradient estimation of implicit distributions \cite{li2018gradient, shi2018spectral}, but requires additional calculation costs. Moreover, it has been shown that functional priors induced by standard Gaussian weight priors exhibit spiking behavior \citep{wenzel2020good, tran2020all}, which can be problematic in training. For these reasons, we adopt the latter approach and impose independent and identical priors on each element of $\mathbf{h}$ for simplicity of implementation. Setting independent priors on features of different data points may seem weird at first, because it does not impose any functional smoothness of feature extractor $h$, unlike typical GP regression \citep{rasmussen2006gaussian}. However, we expect that it still yields sufficient smooth functions owing to a well-known inductive bias in neural network training, leading to generalizing local minima \citep{neyshabur2015search, mandt2017stochastic}.
Further investigation on priors, mainly about the correlation between data \citep{wilson2020bayesian}, is required in future work.

We consider three distributions for priors: normal, Cauchy, and uniform. Cauchy, which is known as a weakly informative prior, is often preferred for robustness as it places less density at the mean owing to the heavier tails \citep{gelman2006prior}. We expect this to be more suitable for feature-space priors because the activation of a trained network often shows heavy-tailed distributions \cite{peluchetti2020stable}.
Uniform prior is improper, and imposing such a prior corresponds to removing the effect of priors in the update rule \eqref{eq8}. We explore their practical performance in \autoref{ablation} and adopt Cauchy as a default prior for its superior performance.

\begin{table*}[htbp]
\caption{Results for Wide ResNet-16-4 on CIFAR-10 with an ensemble size of 10, evaluated over 5 seeds.}
\label{table2}
\vskip 0.15in
\begin{center}
\begin{small}
\begin{sc}
\begin{tabular}{lccccc}
\toprule
Method & Accuracy($\uparrow$) & NLL($\downarrow$)  & Brier($\downarrow$)  & ECE($\downarrow$)  & cA / cNLL / cBrier / cECE\\
\midrule
Single & 95.4 $\pm$ 0.2& 0.145 $\pm$  0.006&  0.069 $\pm$ 0.003 & 0.007 $\pm$ 0.000 & 73.7 / 0.796 / 0.349 / \textbf{0.020}\\
\midrule
Deep Ensembles    & 96.4 $\pm$ 0.1& 0.110 $\pm$ 0.001& 0.054 $\pm$ 0.001 & 0.007 $\pm$ 0.000 &76.7 / 0.698 / 0.310 / 0.025\\
weight-WGD & 96.4 $\pm$ 0.1& 0.111 $\pm$ 0.002& 0.054 $\pm$ 0.001 &0.007 $\pm$ 0.001 &76.7 / 0.702 / 0.312 / 0.026\\
function-WGD    & 96.1 $\pm$ 0.1 & 0.124 $\pm$ 0.001 & 0.059 $\pm$ 0.001  & 0.007 $\pm$ 0.001&75.7 / 0.736 / 0.322 / 0.024\\
feature-WGD    & \textbf{96.5 $\pm$ 0.1}& \textbf{0.107 $\pm$ 0.001} & \textbf{0.052 $\pm$ 0.001} & \textbf{0.006 $\pm$ 0.001}&\textbf{77.3} / \textbf{0.681} / \textbf{0.302} / \textbf{0.020}\\
\bottomrule
\end{tabular}
\end{sc}
\end{small}
\end{center}
\vskip -0.1in
\end{table*}

\begin{table*}[htbp]
\caption{Results for Wide ResNet-16-4 on CIFAR-100 with an ensemble size of 10, evaluated over 5 seeds.}
\label{table1}
\vskip 0.15in
\begin{center}
\begin{small}
\begin{sc}
\begin{tabular}{lccccc}
\toprule
Method & Accuracy($\uparrow$) & NLL($\downarrow$)  & Brier($\downarrow$)  & ECE($\downarrow$)  & cA / cNLL / cBrier / cECE\\
\midrule
Single & 77.4 $\pm$ 0.3& 0.835 $\pm$  0.007&  0.316 $\pm$ 0.003 & 0.030 $\pm$ 0.003 & 46.7 / 2.279 / 0.658 / 0.035\\
\midrule
Deep Ensembles    & 82.3 $\pm$ 0.2& 0.632 $\pm$ 0.004& 0.249 $\pm$ 0.001 & 0.020 $\pm$ 0.001 &52.9 / 1.971 / 0.590 / 0.032\\
weight-WGD & 82.3 $\pm$ 0.1& 0.633 $\pm$ 0.002& 0.250 $\pm$ 0.001 &0.021 $\pm$ 0.001 &52.8 / 1.967 / 0.589 / 0.031\\
function-WGD    & 79.0 $\pm$ 0.1 & 0.715 $\pm$ 0.003 & 0.286 $\pm$ 0.001  & 0.018 $\pm$ 0.002&49.5 / 2.133 / 0.623 / 0.034\\
feature-WGD    & \textbf{82.9 $\pm$ 0.2}& \textbf{0.624 $\pm$ 0.002} & \textbf{0.243 $\pm$ 0.001} & \textbf{0.017 $\pm$ 0.001}&\textbf{53.5} / \textbf{1.955} / \textbf{0.584} / \textbf{0.029}\\
\bottomrule
\end{tabular}
\end{sc}
\end{small}
\end{center}
\vskip -0.1in
\end{table*}

\subsection{Computational overhead}
Compared to naive Deep Ensembles, we require additional computation to calculate the repulsive term between members, as shown in \eqref{eq9}. Denoting the batch size as $B$ and the feature dimension of each network as $H$, we need $O(n^2 B H)$ time to calculate the projection basis $\Psi_r$ by exact SVD. Assuming that the kernel evaluation is $O(r)$ (e.g., RBF), evaluating $\eqref{eq9}$ takes $O(n^2r + rBH)$ time. On the other hand, the back-propagation, which both methods have in common, takes $O(nBD)$ (or $O(BD)$ with model parallelization), where $D$ is the number of weights in the network.
In typical classification models such as ResNet, $H \sim 10^3$, $B \sim 10^2$, and $D \sim 10^7$. Therefore, the overall computational cost is dominated by the back-propagation for a practical ensemble size $n \sim 10$. In our experiment on CIFAR-100 for Wide ResNet-16-4 with an ensemble size of 10, feature-WGD takes approximately 17s for one epoch, while Deep Ensembles take 16s on four A100 GPUs.

\section{Experiments}
\label{experiments}

\begin{table*}[htbp]
\caption{Results for ResNet-50 on ImageNet with an ensemble size of 5. Note that we only evaluate 1 run due to the computational cost.}
\label{imagenet}
\vskip 0.15in
\begin{center}
\begin{small}
\begin{sc}
\begin{tabular}{lccccc}
\toprule
Method & Accuracy($\uparrow$) & NLL($\downarrow$)  & Brier($\downarrow$)  & ECE($\downarrow$)  & cA / cNLL / cBrier / cECE\\
\midrule
Single & 75.7 & 0.954&  0.338 & 0.018 & 37.7 / 3.235 / 0.738 / 0.021\\
\midrule
Deep Ensembles    & \textbf{78.0} & \textbf{0.853} & \textbf{0.309}  & 0.019  &40.9 / 3.011 / 0.706 / \textbf{0.015}\\
feature-WGD    & \textbf{78.0}& {0.859} & \textbf{0.309} & \textbf{0.015} &\textbf{42.4} / \textbf{2.923} / \textbf{0.693} / 0.018\\
\bottomrule
\end{tabular}
\end{sc}
\end{small}
\end{center}
\vskip -0.1in
\end{table*}

In this section, we present the results on popular image classification tasks: CIFAR-10, CIFAR-100 \citep{krizhevsky2009learning}, and ImageNet \citep{deng2009imagenet}. For CIFAR, we use a Wide ResNet-16-4 as the base architecture for its compactness and decent performance \cite{zagoruyko2016wide}. For ImageNet, we use ResNet-50 as it is the most commonly benchmarked model \citep{he2016deep}.
We follow the standard scheduling, augmentation, and regularization schemes in the literature \cite{chen2020online}, which are summarized in \autoref{impl}. Note that these training schemes have been heavily tuned in prior works to prevent overfitting, so it is hard to improve ensemble performance simply by improving single model performance. In addition, it is worth mentioning that we use SGD with Nesterov acceleration as an optimizer, whereas Adam \cite{kingma2015adam} has been traditionally used for particle-based inference of neural networks \cite{liu2016stein, wang2019function, d2021stein}. We think this yields a more fair and practical comparison of particle-based inference methods and Deep Ensembles since SGD generally performs significantly better than Adam on these classification tasks \citep{wilson2017marginal}. 
For a kernel used in WGD, we adopt a standard RBF kernel and determine its bandwidth by the median heuristic \citep{scholkopf2002learning}. Code is available at: \url{https://github.com/DensoITLab/featurePI}.

\textbf{Metrics.} 
We report the test accuracy, negative log-likelihood (NLL), Brier score \cite{brier1950verification}, and the expected calibration error (ECE) \cite{naeini2015obtaining}. For calibration metrics, we follow the procedure from \citep{ashukha2019pitfalls}, which evaluates the metrics after temperature scaling \cite{guo2017calibration}. In addition to the in-domain evaluation, we measure robustness to image perturbations by evaluating these metrics on the corrupted versions of these datasets (CIFAR-10-C, CIFAR-100-C, and ImageNet-C) \cite{hendrycks2019benchmarking}. They apply a set of 15 common visual corruptions with intensities ranging from 1 to 5. We average each metric over all corruption types and intensities (cA(ccuracy), cNLL, cBrier, and cECE).

\subsection{Comparison of WGD on different space and Deep Ensembles}
\label{sec:4.1}

Firstly, we test our central hypothesis, which states that promoting feature space diversity improves ensemble performance. To this end, we compare WGD with varying inference space (\{weight, function, feature\}-WGD) and gold-standard Deep Ensembles, as well as a single model baseline. Except for the difference in the inference space and prior parameters determined by the standard cross-validation, we adopted the same training scheme for all methods.

\textbf{CIFAR-10 and CIFAR-100.} We train Wide ResNet-16-4 with an ensemble size of 10 on both datasets. The result presented in \autoref{table2} (CIFAR-10) and \autoref{table1} (CIFAR-100) show that feature-WGD performs the best in terms of accuracy and calibration metrics on both the in-domain and the corrupted test set. For in-domain evaluation, the improvement of feature-WGD over Deep Ensembles is relatively small on CIFAR-10, because a single model already performs relatively well. On the other hand, a significant performance gain (e.g., $+0.6$\% in accuracy) for the more difficult CIFAR-100 is observed. For corrupted data, we observe much better 
accuracy and calibration of feature-WGD on both datasets. This highlights the robustness induced by the exploitation of various data views. 

The performance of weight-WGD almost coincides with that of Deep Ensembles, indicating that the weight-space repulsive term cannot exploit diversity beyond the difference in initialization. As reported in previous works \citep{d2021repulsive, d2021stein}, function-WGD shows a severe underfitting even in the optimized prior, which can be attributed to the harmfulness of the function-space repulsive term on these datasets. We think this is clear evidence of the advantages of feature-space inference over weight or function-space inference.

\begin{figure}[htbp]
\vskip 0.2in
\begin{center}
\centerline{\includegraphics[width=0.95\columnwidth]{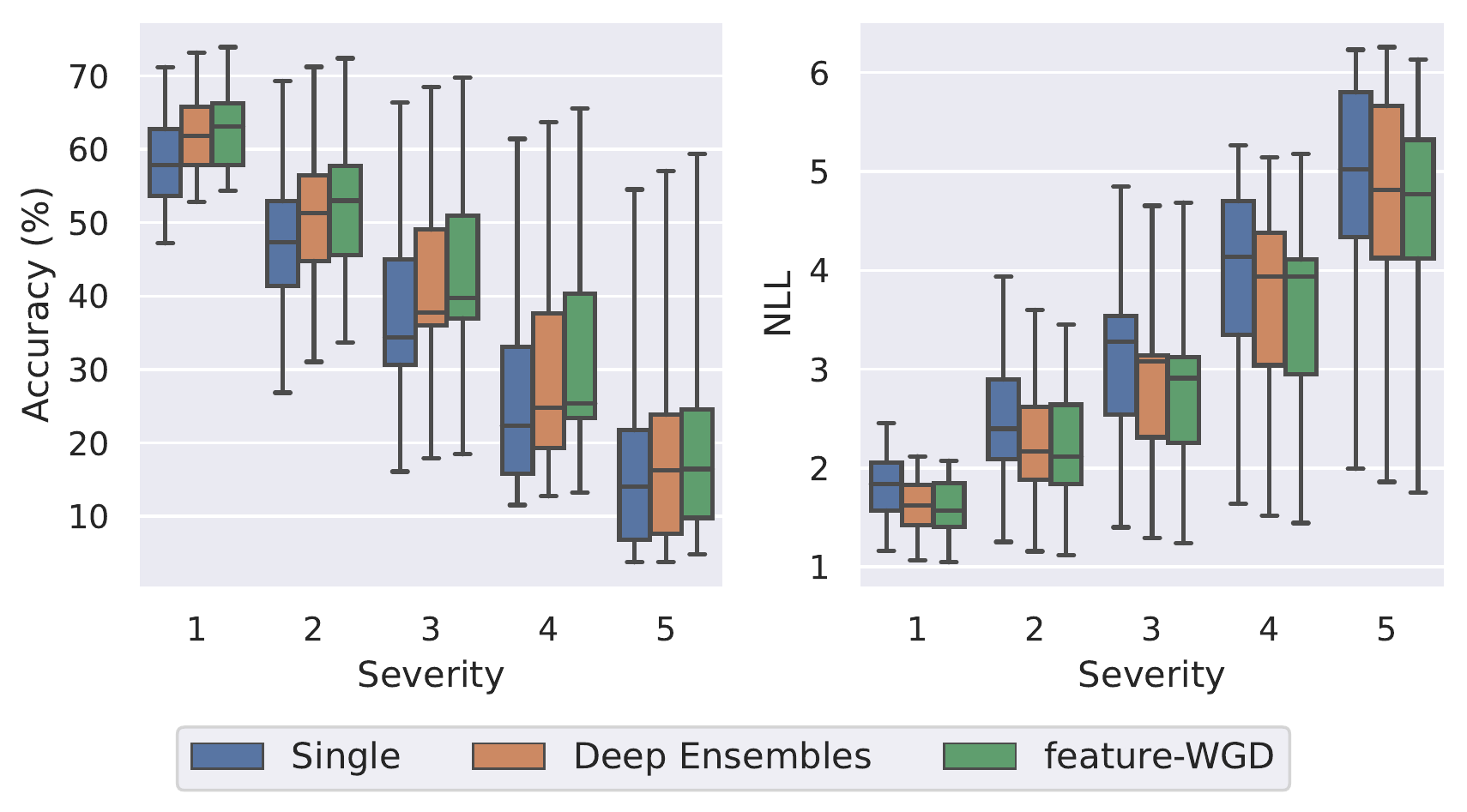}}
\caption{Results for ResNet-50 on ImageNet-C with an ensemble size of 5. We plot accuracy and NLL evaluated on 15 corruption types with corruption severity ranging 1-5.}
\label{imagenet-c}
\end{center}
\vskip -0.2in
\end{figure}

\textbf{ImageNet.} We train ResNet-50 with an ensemble size of 5. Here we only evaluate feature-WGD and Deep Ensembles. The results are presented in \autoref{imagenet}. Although the in-domain accuracy and calibration scores are almost comparable, robustness to corruption is clearly improved (e.g., $+1.5$\% in accuracy) by feature-WGD. 
\autoref{imagenet-c} examines the performance under corruption in more detail by plotting results across corruption types for each corruption severity. Feature-WGD shows better robustness than Deep Ensembles even in the intense corruption.

\subsection{Comparison with SOTA ensemble methods}
\label{sec:4.2}
Here, we compare feature-WGD with other ensemble methods proposed in the literature from a practical viewpoint. We select ADP \citep{pang2019improving} and DICE \citep{rame2021dice} as both report superior performance to Deep Ensembles. Note that we do not compare with Bayesian methods here, because generally they have not shown superior performance to Deep Ensembles with the same ensemble size \citep{ashukha2019pitfalls}. 
Due to code availability, we compare the reported values for ResNet-32 and Wide ResNet-28-2 on CIFAR-100 in \cite{rame2021dice} and follow the same training procedure for feature-WGD.

The results are presented in \autoref{table5}. For Wide ResNet-28-2 with an ensemble size of 3, feature-WGD performs the best, as is DICE. For ResNet-32 with an ensemble size of 4, feature-WGD outperforms Deep Ensembles and ADP but falls short of DICE. We think one possible explanation for this is the insufficient width of the networks, as discussed later in \autoref{ablation}.
We note that DICE requires additional networks to estimate mutual information between models and adversarial training on them, whereas feature-WGD only modifies the gradient calculation in standard training. Overall, these results suggest that our method works well as a practical ensemble method, even though those ensemble size is rather small for a Bayesian method.

\begin{table}[htbp]
\caption{Comparison with SOTA ensemble methods on CIFAR-100. We include the values reported in \cite{rame2021dice}. Evaluated over 5 seeds.}
\label{table5}
\vskip 0.15in
\begin{center}
\begin{small}
\begin{sc}
\begin{tabular}{lcccr}
\toprule
 Method & ResNet-32$\times$4 & WRN-28-2$\times$3  \\
\midrule
Deep Ensembles     & 77.4 $\pm$ 0.1& 80.0 $\pm$ 0.2&\\
ADP & 77.5 $\pm$ 0.3& 80.0 $\pm$ 0.2&\\
DICE    & \textbf{77.9 $\pm$ 0.1} & \textbf{80.6 $\pm$ 0.1}&\\
feature-WGD    & 77.6 $\pm$ 0.2 & \textbf{80.6 $\pm$ 0.2}&\\
\bottomrule
\end{tabular}
\end{sc}
\end{small}
\end{center}
\vskip -0.1in
\end{table}

\subsection{In-depth experiments}

\label{ablation}

This section delves deeper into the nature of our feature space particle inference and presents some guidelines about the choice of prior and base architecture for practical implementations.

\begin{figure}[htbp]
\vskip 0.2in
\begin{center}
\centerline{\includegraphics[width=\columnwidth]{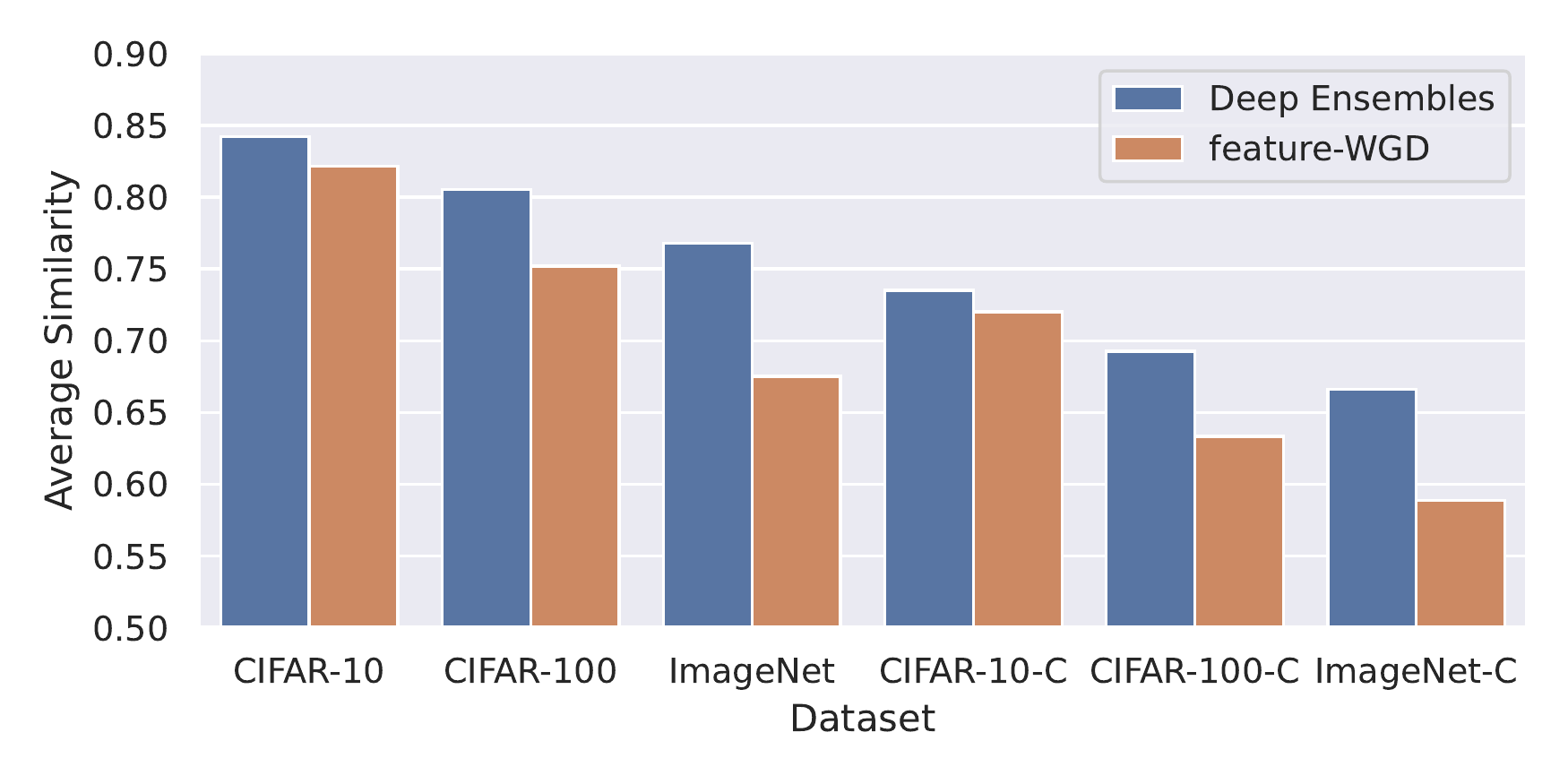}}
\caption{Average similarities of class activation maps produced by Grad-CAM among ensemble members of Deep Ensembles and feature-WGD on CIFAR-10, CIFAR-100, ImageNet, and those corrupted counterparts.}
\label{cam_sim}
\end{center}
\vskip -0.2in
\end{figure}

\textbf{Quantitative evaluation of feature diversity.} Here we investigate how diverse the features obtained in feature-WGD are when compared to Deep Ensembles. We examine the diversity of class activation maps created by Grad-CAM \cite{selvaraju2017grad} instead of raw feature values because feature space is not shared among members in Deep Ensembles. Specifically, we compute an inner product of $l_2$-normalized class activation maps of an input image between two ensemble members as a similarity measure and average it across all pairs in an ensemble and test images. 
The results in \autoref{cam_sim} show that the similarity of feature-WGD is consistently lower than that of Deep Ensembles across datasets used in \autoref{sec:4.1}, showing that members in the feature-WGD ensemble look at more various parts of images from each other. Furthermore, a larger drop in similarity is observed in more complex, multi-object datasets (ImageNet) than in less complicated, object-centric datasets (CIFAR-10). For corrupted datasets, where some features might be unavailable to classify data, we observe the similarity of both methods decreases from non-corrupted datasets, and feature-WGD further decreases it. From this and results in \autoref{sec:4.1}, we can see that feature-WGD indeed exploits the multi-view structure of data and achieves higher classification robustness.

\textbf{Prior choice.} To examine the practical choice of priors in feature-WGD, we experiment with our CIFAR-100 setup for three priors: normal, Cauchy, and uniform. Prior parameters are optimized by cross-validation. The results in \autoref{table4} show that Cauchy performs the best both in terms of accuracy and calibration, which may be attributed to the heavy-tail nature of neural network features, as discussed in \autoref{prior}. The uniform prior performs worse than Deep Ensembles, indicating that some informative priors are necessary for feature-WGD. The harmfulness of uniform priors has been suggested in the Bayesian statistics literature \citep{gelman2020holes}.
\begin{table}[htbp]
\caption{Results for Wide ResNet-16-4 on CIFAR-100 with an ensemble size of 10 trained with feature-WGD on different priors. Averaged over 5 seeds.}
\label{table4}
\vskip 0.15in
\begin{center}
\begin{small}
\begin{sc}
\begin{tabular}{lcccc}
\toprule
 Prior & Accuracy & NLL & Brier & ECE  \\
\midrule
Normal     & 82.7 & 0.628 & 0.245 & \textbf{0.017}\\
Cauchy & \textbf{82.9} & \textbf{0.624} & \textbf{0.243} & \textbf{0.017}\\
Uniform    & 82.1 & 0.634 & 0.251 & 0.020\\
\bottomrule
\end{tabular}
\end{sc}
\end{small}
\end{center}
\vskip -0.1in
\end{table}

\begin{figure}[htbp]
\vskip 0.2in
\begin{center}
\centerline{\includegraphics[width=\columnwidth]{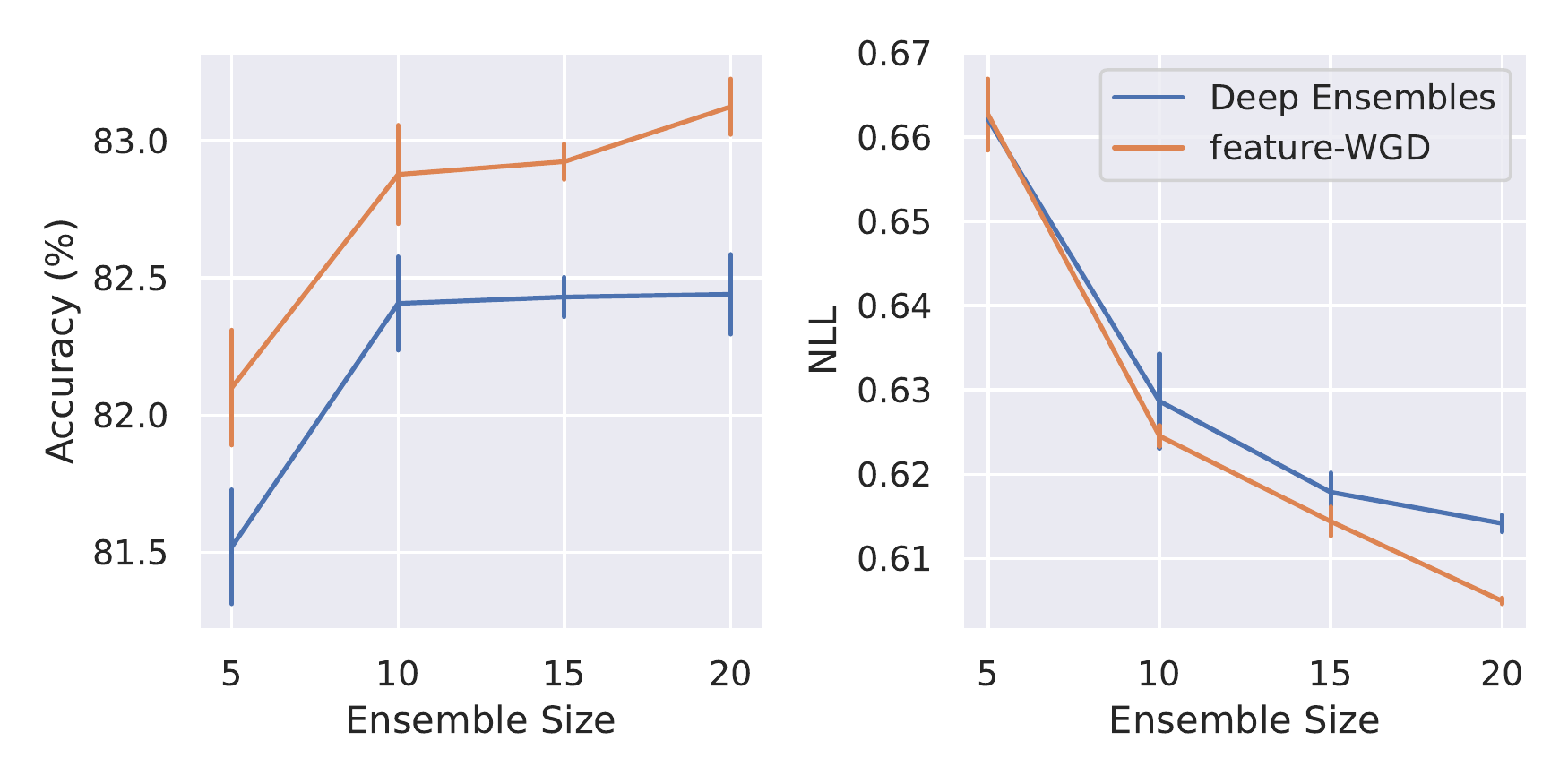}}
\caption{Accuracies (left) and NLLs (right) for Wide ResNet-16-4 on CIFAR-100 in the large ensemble sizes, evaluated over 3 seeds.}
\label{large_size}
\end{center}
\vskip -0.2in
\end{figure}

\textbf{Improvement in larger ensemble size.} Thus far, feature-WGD demonstrates better performance than Deep Ensembles with the same ensemble size. Then the natural question is, {\it when increasing an ensemble size, does the performance of Deep Ensembles catch up with feature-WGD?}  \autoref{large_size} shows test accuracy and negative log-likelihood of these methods when increasing an ensemble size up to 20. We can observe Deep Ensembles face saturation in both metrics, whereas feature-WGD improves its performance even in large ensemble sizes.
This possibly indicates the limitations of relying only on differences in initial values to find a variety of solutions and the benefit of explicitly promoting feature diversity.

\textbf{Need for network width.} As we promote diversity in shared feature space, we expect feature-WGD needs more redundant feature capacity than standard training to exploit multi-view structures. \autoref{table6} shows the accuracy results of feature-WGD and Deep Ensembles for our CIFAR-100 setting with varying network width factors. As expected, the improvement of accuracy by feature-WGD is relatively small in narrow networks (width factor 2) compared to wider networks (width factors 4 and 8). We believe this is one reason for the relatively poor performance of feature-WGD for ResNet-32 in \autoref{table5}, which has only 64 feature dimensions to classify 100 classes.

\begin{table}[htbp]
\caption{Accuracy improvement of feature-WGD over Deep Ensembles on CIFAR-100 with an ensemble size of 10 for different network width factors (Wide ResNet-16-\{2, 4, 8\}). Averaged over 3 seeds.}
\label{table6}
\vskip 0.15in
\begin{center}
\begin{small}
\begin{sc}
\begin{tabular}{lcccc}
\toprule
Width Factor & 2 & 4    & 8 \\
\midrule
Deep Ensembles    & 80.1 & 82.3   & 83.6  \\
feature-WGD & 80.3 & 82.9   & 84.1  \\
\midrule
Improvement   & $+$0.2 & $+$0.6 &  $+$0.5  \\
\bottomrule
\end{tabular}
\end{sc}
\end{small}
\end{center}
\vskip -0.1in
\end{table}

\section{Related Work}
In addition to related works addressed in \autoref{background}, we highlight some relevant studies on ensembles and feature learning of neural networks. 

\textbf{Bayesian neural networks.}
Traditional MCMC methods \citep{mackay1992practical, neal1996bayesian} are considered as the gold standard for BNNs \citep{wenzel2020good}, but face difficulties in modern large-scale learning. Many practical BNNs are proposed, including variational inference \citep{kingma2015variational, wen2018flipout}, K-FAC Laplace approximation \citep{ritter2018scalable}, SWAG \citep{maddox2019simple}, subspace inference \citep{izmailov2020subspace}, and SG-MCMC \citep{welling2011bayesian, zhang2020cyclical}. These works enable Bayesian inference on large problems, but they still fall short of Deep Ensembles in terms of both accuracy and uncertainty estimation \cite{ashukha2019pitfalls}.

\textbf{Non-Bayesian ensemble methods.}
Beyond random initialization \cite{lakshminarayanan2017simple}, some studies apply different augmentations \citep{dvornik2019diversity} or hyperparameters \citep{wenzel2020hyper} to improve the diversity, but they are highly dependent on domain knowledge and engineering. For methods that explicitly diversify network outputs, ADP \citep{pang2019improving} promotes the orthogonality of the non-maximal predictions.
DICE \citep{rame2021dice} shares some concepts with our work in that it promotes network diversity by reducing feature correlations. They use additional networks to estimate mutual information and adversarial training, incurring extra training costs.

\textbf{Spurious correlations of features.}
It is generally known that trained neural networks frequently rely on features that are predictive of the target in the training data, but irrelevant to the underlying true labeling function (e.g., backgrounds co-occurring with foreground objects) \cite{geirhos2020shortcut}. Because depending solely on these spurious features impairs generalization performance, several methods are proposed to avoid them \cite{arjovsky2019invariant, kirichenko2022dfr, pagliardini2022agree}. For example, the distributionally robust optimization (DRO) framework \cite{sagawa2020distributionally} optimizes the worst-case loss over a set of pre-specified groups in the training data instead of the averaged loss to obtain invariant features across groups. In this light, our ensemble method addresses a similar issue by gathering plausible (both spurious and non-spurious) features that explain training data and improves generalization performance even in the presence of distributional corruptions. Note that these studies generally assume that input data from possible test distributions (or out of distributions) are accessible in training, which we do not assume in this study.

\section{Conclusion}
We have introduced a feature-space particle inference framework for neural network ensembles, which encourages each ensemble member to exploit various data views. Our extensive experiments have shown that feature-space inference significantly outperforms traditional Deep Ensembles and weight/function-space inference in terms of accuracy, uncertainty estimation, and robustness. 

Although we consider only a final linear layer as a shared classifier in this work, extending shared parts to shallower layers may be engaging in the future. This may suggest which level of feature variation contributes to ensemble performance.
For more efficient implementations, we can use the rank-1 parametrization of an ensemble \cite{wen2020batchensemble, dusenberry2020efficient} as a backbone feature extractor. Reducing the evaluation of the repulsive term is also promising, especially for distributed training.
Furthermore, we can readily apply the recently developed techniques of particle inference \cite{gallego2018stochastic, zhang2020stochastic} for further performance gains.

\section*{Acknowledgements}
We would like to thank Mitsuru Ambai and Yuichi Yoshida for helpful feedback on the paper.

\bibliography{main}
\bibliographystyle{icml2022}

\onecolumn
\appendix

{\Huge \bf Appendix}

\section{Algorithm}
Here we outline the detailed procedure of feature-WGD, possibly under model parallel execution, in Algorithm \autoref{alg:feature}. Although feature-WGD requires additional communication cost ($\mathtt{MPI\_Allgather}$) to calculate interactions between ensemble members, communicated variables are not so high-dimensional (at most \{feature dimension\} $\times$ \{batch size\}) compared to the gradient communication in standard data parallel executions, which requires parameter dimension communications. Empirically, these overheads are negligible in practical ensemble size ($\sim 10$).
\label{algorithm}
\begin{algorithm}[htbp]
   \caption{feature-WGD (in parallel)}
   \label{alg:feature}
\begin{algorithmic}[1]
   \STATE {\bfseries Input:} training data $\mathcal{D}$, the number of optimization steps $T$, step size $\{\alpha_t\}_{t=1}^T$, weight decay parameter $\lambda$, projection dimension $r$
   \STATE {\bfseries Output:} optimized parameters $\{w_i\}_{i=1}^n, \theta.$
   \STATE Initialize parameters $\{w_i\}_{i=1}^n, \theta.$
   \FOR{$t=1$ {\bfseries to} $T$}
   \STATE Draw a mini-batch $\{x_b, y_b\}_{b=1}^B \sim \mathcal{D}.$
   \FOR{$i=1$ {\bfseries to} $n$}
   \STATE Construct a feature vector through feed-forwarding: $$\mathbf{h}_i = \mathrm{vec}(\{h(x_b; w)\}_b).$$
   \STATE Calculate gradients: $$\mathbf{g}^{\mathrm{data}}_{i} = \nabla_{\mathbf{h}_i} \log p(\{y_b\}_b | \mathbf{h}_i),$$ $$\mathbf{g}^{\mathrm{prior}}_{i} = \nabla_{\mathbf{h}_i} \log p(\mathbf{h}_i).$$
   \ENDFOR
   \STATE (Perform $\mathtt{MPI\_Allgather}$ for $\left\{\mathbf{g}^{\mathrm{data}}_{i}\right\}_{i=1}^n.$)
   \STATE Construct basis: $$\Psi, \Sigma, V = \mathrm{SVD([\mathbf{g}^{\mathrm{data}}_{1}, \mathbf{g}^{\mathrm{data}}_{2},\ldots, \mathbf{g}^{\mathrm{data}}_{n}])},$$ $$\Psi_r= \Psi[:, :r].$$
   \FOR{$i=1$ {\bfseries to} $n$}
   \STATE Project a feature vector: $$\mathbf{z}_i = \Psi_r^\top \mathbf{h}_i. $$
   \ENDFOR
   \STATE (Perform $\mathtt{MPI\_Allgather}$ for $\left\{\mathbf{z}_{i}\right\}_{i=1}^n.$)
   \FOR{$i=1$ {\bfseries to} $n$}
   \STATE Calculate a feature update direction: $$v_i^{\mathbf{h}} = \mathbf{g}^{\mathrm{data}}_{i} +  \mathbf{g}^{\mathrm{prior}}_{i} 
    -\Psi_r \frac{\sum_{j=1}^n \nabla_{\mathbf{z}_i} k( \mathbf{z}_i, \mathbf{z}_j)}{\sum_{j=1}^n k(\mathbf{z}_i, \mathbf{z}_j)}.$$
    \STATE Calculate a parameter update direction through back-propagation: $$v_i^w = \frac{1}{B} \left(\frac{\partial \mathbf{h}_i}{\partial w_i}\right)^\top v_i^\mathbf{h} - \lambda w_i,$$
    $$ v_i^\theta = \frac{1}{B}   \sum_{b=1}^B \nabla_{\theta} \log p\left(y_b \middle| c (h (x_b; w_i);\theta)\right) - \lambda \theta. $$ 
    \STATE Update feature extractor parameters: $$ w_i \leftarrow w_i + \alpha_t v_i^w,$$
   \ENDFOR
   \STATE (Perform $\mathtt{MPI\_Allgather}$ for $\left\{ v_i^\theta \right\}_{i=1}^n.$)
   \STATE Update classifier parameters: 
   $$\theta \leftarrow \theta + \frac{\alpha_t}{n} \sum_{i=1}^n  v_i^\theta. $$
   \ENDFOR
\end{algorithmic}
\end{algorithm}

\section{Implementation Details}
\label{impl}

\subsection{Training schemes}
Classical hyperparameters in the literature are taken from \citep{chen2020online}. We use stochastic gradient descent with Nesterov momentum for optimization, and adopt standard augmentation schemes (random crop and horizontal flip). Note that these hyperparameter are common to all methods including Deep Ensembles and \{weight, function, feature\}-WGD.
\begin{table}[htbp]
\caption{Hyperparameter values for training on CIFAR-10, CIFAR-100, and ImageNet.}
\vskip 0.15in
\begin{center}
\begin{small}
\begin{sc}
\begin{tabular}{lcccc}
\toprule
Dataset & CIFAR-10 & CIFAR-100 & ImageNet \\
\midrule
epoch  & $300$ & $300$ & $90$\\
batch size  & $128$ & $128$ & $256$\\
base learning rate  &  $0.1$ & $0.1$ & $0.1$\\
lr decay ratio  &  $0.1$ & $0.1$ & $0.1$\\
lr decay epochs  &  $[150, 225]$ & $[150, 225, 250]$ & $[30, 60]$\\
momemtum & 0.9 & 0.9 & 0.9 \\
weight decay  &  $5 \times 10^{-4}$ & $5 \times 10^{-4}$ & $1 \times 10^{-4}$\\
\bottomrule
\end{tabular}
\end{sc}
\end{small}
\end{center}
\vskip -0.1in
\end{table}

\subsection{Prior parameters}
Because feature values after global average pooling are positive in ResNets, we consider priors (normal, Cauchy, and uniform) to be supported only on the positive parts as follows:
\begin{align}
&\mathrm{HalfNormal}(x; \sigma) =  \begin{cases}
\sqrt{\frac{2}{\pi \sigma^2}}\mathrm{e}^{-x^2/2\sigma^2} & x \geq 0, \\
0 & x < 0
\end{cases}\\
&\mathrm{HalfCauchy}(x; \sigma) =  \begin{cases}
\frac{2}{\pi \sigma} \frac{1}{1+x^2/\sigma^2} & x \geq 0, \\
0 & x < 0.
\end{cases}
\end{align}

In \autoref{ap1}-\ref{ap2}, we present prior parameters on each experiment in \autoref{experiments}.

\begin{table}[htbp]
\caption{Prior parameters for Wide ResNet-16-4 on CIFAR-10 in \autoref{table2}.}
\label{ap1}
\vskip 0.15in
\begin{center}
\begin{small}
\begin{sc}
\begin{tabular}{lcccc}
\toprule
Method & weight-WGD & function-WGD & feature-WGD \\
\midrule
prior  & normal & Cauchy & Cauchy\\
prior scale $1 / \sigma^2$  & $1 \times 10^{-3}$ & $1 \times 10^{-6}$ & $ 1 \times 10^{-3}$\\
projection dim $r$ & $5$ & $5$ & $5$\\
\bottomrule
\end{tabular}
\end{sc}
\end{small}
\end{center}
\vskip -0.1in
\end{table}

\begin{table}[htbp]
\caption{Prior parameters for Wide ResNet-16-4 on CIFAR-100 in \autoref{table1}.}
\vskip 0.15in
\begin{center}
\begin{small}
\begin{sc}
\begin{tabular}{lcccc}
\toprule
Method & weight-WGD & function-WGD & feature-WGD \\
\midrule
prior  & normal & Cauchy & Cauchy\\
prior scale $1 / \sigma^2$  & $1 \times 10^{-3}$ & $1 \times 10^{-6}$ & $ 5 \times 10^{-3}$\\
projection dim $r$  & $5$ & $5$ & $5$\\
\bottomrule
\end{tabular}
\end{sc}
\end{small}
\end{center}
\vskip -0.1in
\end{table}

\begin{table}[htbp]
\caption{Prior parameters for ResNet-50 on ImageNet in \autoref{imagenet}.}
\vskip 0.15in
\begin{center}
\begin{small}
\begin{sc}
\begin{tabular}{lcccc}
\toprule
Method  & feature-WGD \\
\midrule
prior   & Cauchy\\
prior scale $1 / \sigma^2$  &  $2 \times 10^{-3}$\\
projection dim $r$ & $5$\\
\bottomrule
\end{tabular}
\end{sc}
\end{small}
\end{center}
\vskip -0.1in
\end{table}

\begin{table}[htbp]
\caption{Prior parameters for ResNet-32 and Wide ResNet-28-2 on CIFAR-100 in \autoref{table5}.}
\vskip 0.15in
\begin{center}
\begin{small}
\begin{sc}
\begin{tabular}{lcccc}
\toprule
architecture & ResNet-32$\times$4 & WRN-28-2$\times$3  \\
\midrule
prior  & Cauchy & Cauchy \\
prior scale $1 / \sigma^2$  & $1 \times 10^{-3}$ & $1 \times 10^{-3}$\\
projection dim $r$ & $4$ & $3$ \\
\bottomrule
\end{tabular}
\end{sc}
\end{small}
\end{center}
\vskip -0.1in
\end{table}

\begin{table}[htbp]
\caption{Prior parameters for Wide ResNet-16-4 on CIFAR-100 with various priors in \autoref{table4}.}
\label{ap2}
\vskip 0.15in
\begin{center}
\begin{small}
\begin{sc}
\begin{tabular}{lcccc}
\toprule
Prior & normal & Cauchy & uniform \\
\midrule
prior scale $1 / \sigma^2$ & $ 5 \times 10^{-3}$ & $ 5 \times 10^{-3}$ & $-$\\
projection dim $r$ & $5$ & $5$ & $5$\\
\bottomrule
\end{tabular}
\end{sc}
\end{small}
\end{center}
\vskip -0.1in
\end{table}

\section{Additional Experiments}

{\bf Effect of ensembling.} To separate the regularization effect induced by the prior on features from the improvement of feature-WGD, we compare a single model trained with the feature prior term (feature-MAP) to one trained without it (Single). The results summarized in \autoref{table3} show that imposing prior itself does not improve calibration scores on a single model, whereas accuracy is slightly improved. From this, we can see that the superior performance of feature-WGD to Deep Ensembles comes not from the regularization scheme on each model but ensembling and interaction between models.
\begin{table}[htbp]
\caption{Results for single Wide ResNet-16-4 on CIFAR-100, trained with (feature-MAP) and without (Single) the feature prior term. Averaged over 5 seeds.}
\label{table3}
\vskip 0.15in
\begin{center}
\begin{small}
\begin{sc}
\begin{tabular}{lcccc}
\toprule
Method & Accuracy & NLL & Brier & ECE  \\
\midrule
Single & 77.4 & 0.835 & 0.316 & 0.030\\
feature-MAP    & 77.6 & 0.883 & 0.318 & 0.044\\
\bottomrule
\end{tabular}
\end{sc}
\end{small}
\end{center}
\vskip -0.1in
\end{table}

{\bf Ablation study on the projection dimension parameter $r$.} Here we show the influence of the projection dimension parameter $r$ on CIFAR-100 setting in \autoref{sec:4.1} in \autoref{table11}.
\begin{table}[htbp]
\caption{Ablation study on the projection dimension $r$ for Wide ResNet-16-4 on CIFAR-100. Averaged over 3 seeds.}
\label{table11}
\vskip 0.15in
\begin{center}
\begin{small}
\begin{sc}
\begin{tabular}{lcccc}
\toprule
 & $r=3$ & $r=5$ & $r=10$ & no projection\\
\midrule
Accuracy & $82.7$ & $82.9$ & $82.7$ & $82.7$ \\
\bottomrule
\end{tabular}
\end{sc}
\end{small}
\end{center}
\vskip -0.1in
\end{table}

\newpage

\section{Additional figures}
\begin{figure}[htbp]
\vskip 0.2in
\begin{center}
\centerline{\includegraphics[width=0.5\columnwidth]{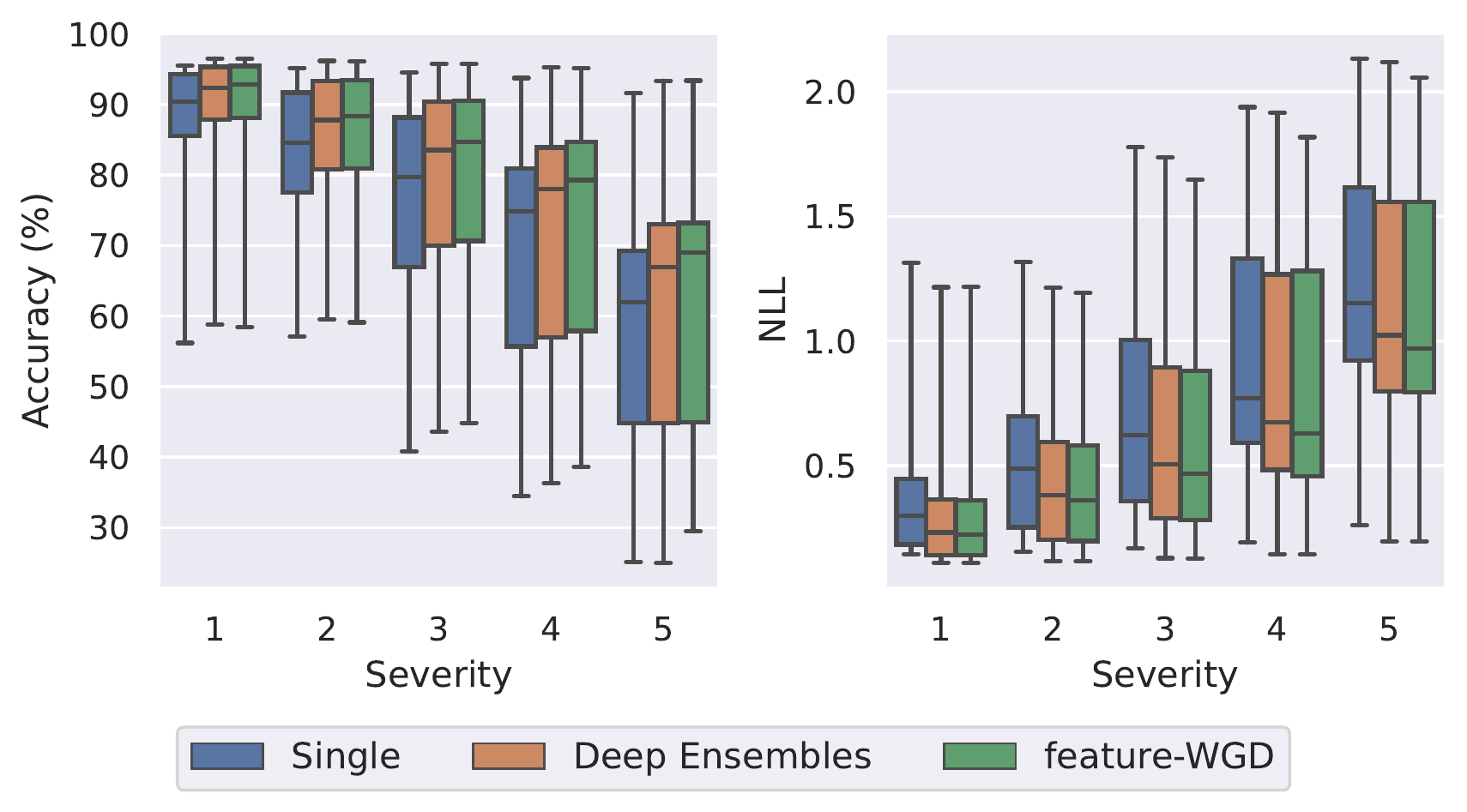}}
\caption{Results for Wide ResNet-16-4 on CIFAR-10-C with an ensemble size of 5. We plot accuracy and NLL evaluated on 15 corruption types for varying corruption severity 1-5.}
\label{cifar10-c}
\end{center}
\vskip -0.2in
\end{figure}

\begin{figure}[htbp]
\vskip 0.2in
\begin{center}
\centerline{\includegraphics[width=0.5\columnwidth]{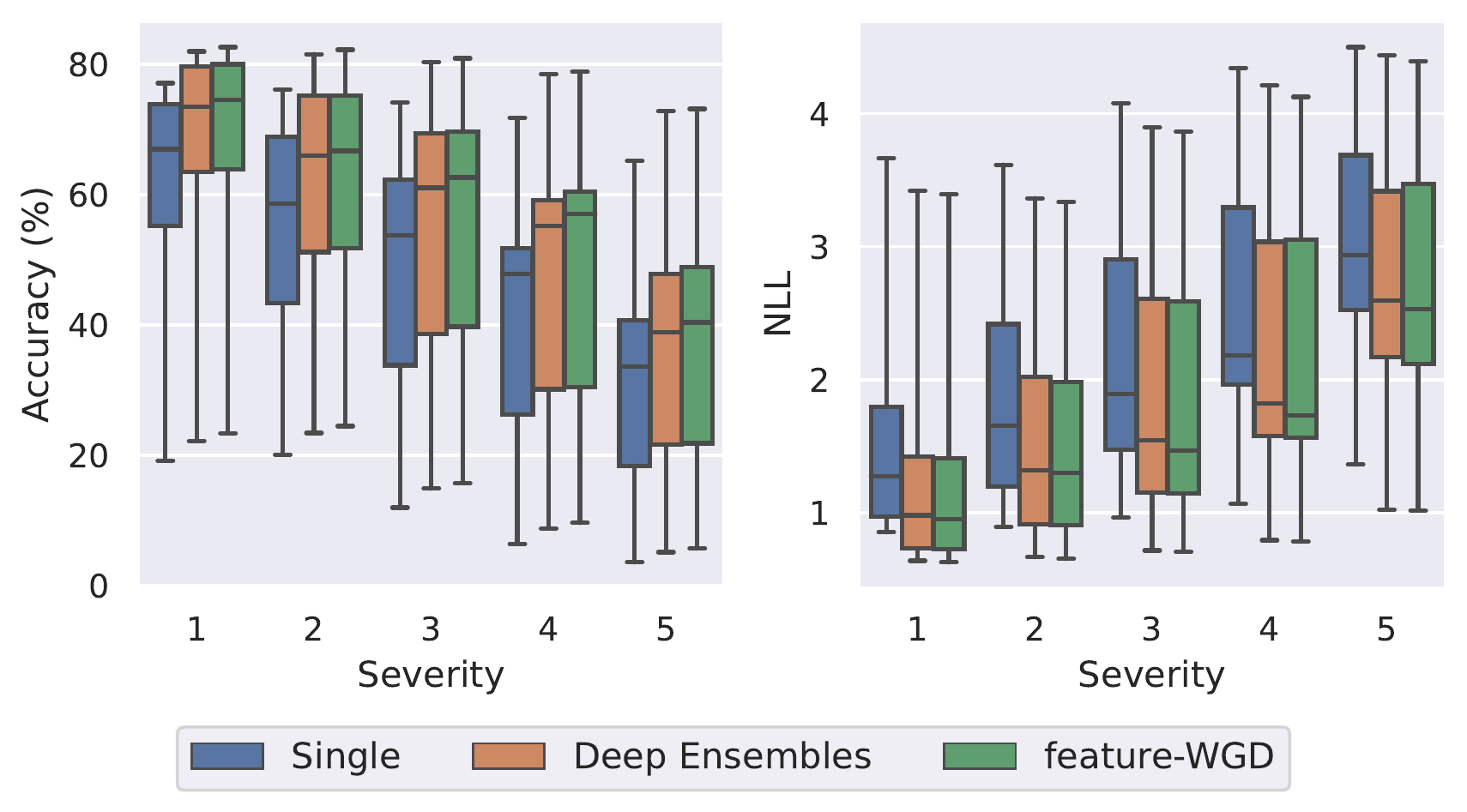}}
\caption{Results for Wide ResNet-16-4 on CIFAR-100-C with an ensemble size of 5. We plot accuracy and NLL evaluated on 15 corruption types for varying corruption severity 1-5.}
\label{cifar100-c}
\end{center}
\vskip -0.2in
\end{figure}

We put results for CIFAR-10-C (\autoref{cifar10-c})  and CIFAR-100-C (\autoref{cifar100-c}) in \autoref{sec:4.1} across corruption types for each corruption severity.

\end{document}